\pdfoutput=1 
\documentclass[conference]{ieeeconf}
\IEEEoverridecommandlockouts
\usepackage{cite}
\usepackage{amsmath,amssymb,amsfonts}
\usepackage{gensymb}
\usepackage{algorithmic}
\usepackage{graphicx}
\usepackage{textcomp}
\usepackage{xcolor}
 \usepackage[T1]{fontenc} 
\usepackage[papersize={8.5in,11in},margin=0.75in,footskip=0.25in, top=1in]{geometry}
\def\BibTeX{{\rm B\kern-.05em{\sc i\kern-.025em b}\kern-.08em
    T\kern-.1667em\lower.7ex\hbox{E}\kern-.125emX}}
\begin{document}

\newcommand{\myparagraph}[1]{\vspace{0.05in}\noindent\textbf{#1}}

\title{\LARGE \bf
Design of a Fully Actuated Robotic Hand With Multiple Gelsight Tactile Sensors
}

\author{Achu Wilson$^{1}$, Shaoxiong Wang, Branden Romero, Edward Adelson 
\thanks{*This work was supported by Toyota Research Institute}
\thanks{$^{1}$ All the authors are with the Computer Science and Artificial Intelligence Lab at Massachusetts Institute of Technology, USA.
               {\tt\small \{achuwils,wang\_sx,brromero\}@mit.edu, adelson@csail.mit.edu}}%
}

\maketitle

\begin{abstract}
This work details the design of a novel two finger robot gripper with multiple Gelsight based optical-tactile sensors covering the inner surface of the hand. The multiple Gelsight sensors can gather the surface topology of the object from multiple views simultaneously as well as can track the shear and tensile stress. In addition, other sensing modalities enable the hand to gather the thermal, acoustic and vibration information from the object being grasped. The force controlled gripper is fully actuated so that it can be used for various grasp configurations and can also be used for in-hand manipulation tasks. Here we present the design of such a gripper.
\end{abstract}

\section{Introduction}
 Tactile sensing makes humans manipulation extremely capable, but is often underrated in robots. Human hands, with their tactile sensing skin, which almost covers the entire surface gives a rich set of data for use in manipulation. It enables humans to get a good sense of an object with a single grasp, whereas in robot hands, the tactile sensing is often confined to a small sensor patch on the finger and would require multiple attempts. In addition to the sense of touch, various other sensory information also helps humans greatly in manipulation, like proprioception, thermal, vibration and acoustic signals generated during physical contact and visual sensing. This provides redundancy
as well as fully comprehensive information of the event \cite{VENKADESAN20071653}. For
example, pressing a mechanically latching switch on a
device provides far better user experience than just a touch sensitive switch.
The mechanical switch provides physical movement, tactile
feedback, a click sound and some visual feedback. Without such multi-sensorial feedback, humans have less confidence on the
consequence of their actions which may eventually compromise
the device usability \cite{Park:2011:TED:2037373.2037376,Koskinen:2008:FTF:1452392.1452453}. This work presents the design of a robot hand, which has been designed with a goal to maximize the tactile sensing area inside the hand. In addition, various other sensor modalities are added to aid in better manipulation capabilities.

\section{Related work}

Gelsight is an optical-tactile sensing technology which can provide rich spatial data about the contact surface. The original Gelsight sensor \cite{5206534} was designed to capture the 3D topography of the contact surface with very high spatial resolution, in the range of micrometers. Since then, it has captured the attraction of roboticists who would love to make use of such a high resolution tactile data in manipulation. Li \textit{et al.} designed a cuboid fingertip sensor which was then used to insert a USB connector by tracking the tactile imprint of the characteristic USB logo on the connector \cite {6943123}. Izzat \textit{et al.} fused the 3D shape generated with Gelsight sensors with global pointcloud to make better sense of the pose of the object, especially when it is occluded \cite{7989460}. Siyuan \textit{et al.} designed an improved Gelsight sensor \cite{8202149} with better geometric accuracy. Later, a compact finger inspired by the Gelsight sensing technique - Gelslim, has been developed \cite{8593661} which are attached to the fingers of a parallel gripper to aid manipulation tasks.

In addition to tactile sensing, approaches to multi-modal sensing for robot manipulation exploring  thermal, auditory and visual sensing\cite{8594169},\cite{8314701} has been proven to improve the performance of manipulation. Sensors like Biotac provides force, vibration and thermal information of the contact surface\cite{6290741}.

\begin{figure}[htp]
    \centering
    \includegraphics[width=9cm]{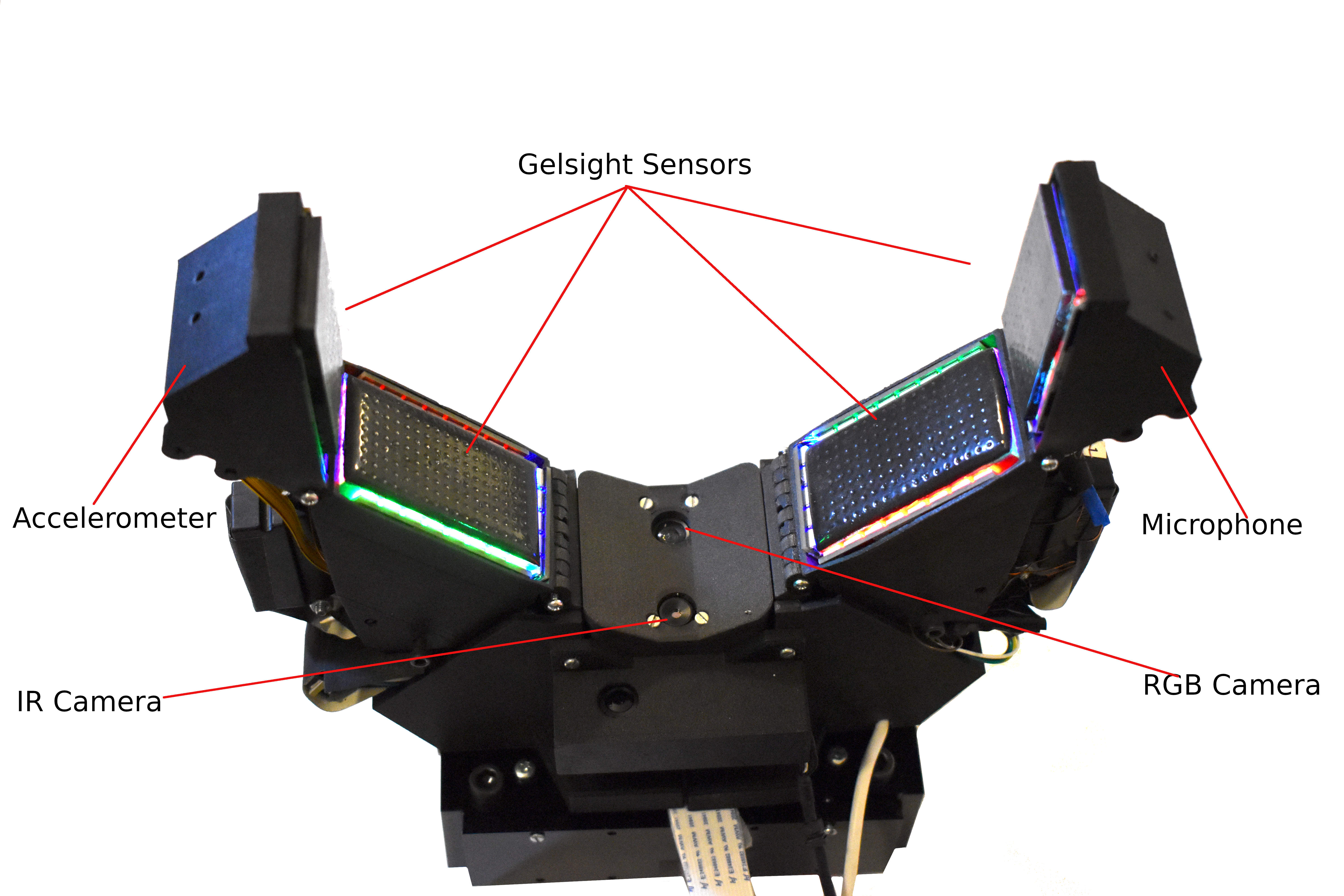}
    \caption{ Gripper with 4 Gelsight Sensors}
    \label{fig:gripper}
\end{figure}

\section{High Level Design}
The main design objectives of the new hand are twofold: first is to have multiple Gelsight based tactile sensors on the same hand and the second is to have it fully actuated. This will, in addition maximize the dexterity so that it can perform in-hand manipulation tasks and and has improved grasp quality, while keeping it compact and cheap. Unlike the earlier approaches which concentrated on fabrication of the sensor, which has to be attached to an external robot gripper, the goal of this work is to come up with a fully actuated gripper. The hand was designed as two identical fingers attached to a base frame, with each finger module containing the sensors and actuators while the base frame house the supporting electronics. The high level design goals are outlined in the following sections

\subsection{Sensing}
The foremost design objective of the hand is to have a gripper with multiple Gelsight optical tactile sensors, covering the maximum inner surface of the gripper. All the previous approaches to tactile sensing and manipulation using Gelsight sensors was confined to the usage of a single sensor. The usage of multiple Gelsight sensors would help the robot to gather much more information simultaneously, from different positions of the object, unlike a single Gelsight sensor with which such data could only be obtained by multiple passes, tactile exploration and servoing. The inner surfaces of each of the four phalanges are
equipped with optical tactile sensors based on Gelsight
technology, designed such that about 85\% of the total
inner surface area of the finger is covered with the tactile sensor,
ensuring the maximum possible amount of tactile information is captured. Design choices were made to maximize the surface area, while keeping the hand compact. In addition to the Gelsight, a multitude of other sensing modalities, which can aid in manipulation are identified prior to design and are integrated into the hand. These includes proprioception, non contact thermal sensing, acoustic and vibratory information from the object being grasped. 

\subsection{kinematics}
A lot of research on underactuated  as well as fully actuated robot grippers has been done in the past. The underactuated hands can adaptively conform to the shape of the objects without the need of sensors or feedback systems, while the fully actuated counterparts would need sensors and feedback mechanisms to do the same. But the fully actuated hands have increased dexterity/manipulability that would allow for better in hand manipulation operations. Most of the fully actuated robot hands are modelled after the human hand and the control of them is quite complex. 
The kinematic design of the hand is chosen such that the it is fully actuated, yet not complex as an anthrophomorphic hand. The fully actuation also helps it to switch modes between different grasping modes like the parallel pinch grasp or an adaptive encapsulating grasp. The two fingers consists of two phalanges each, the lengths of which are calculated for optimum caging grasps as in \cite {8793465} . The ratio of palm length to proximal phalanx to distal phalanx is 1:1.2:0.9
In order to reduce the mechanical, actuation and control complexity, the in-hand manipulation capabilities are restricted to a single plane.

\begin{figure}[htp]
    \centering
    \includegraphics[width=9cm]{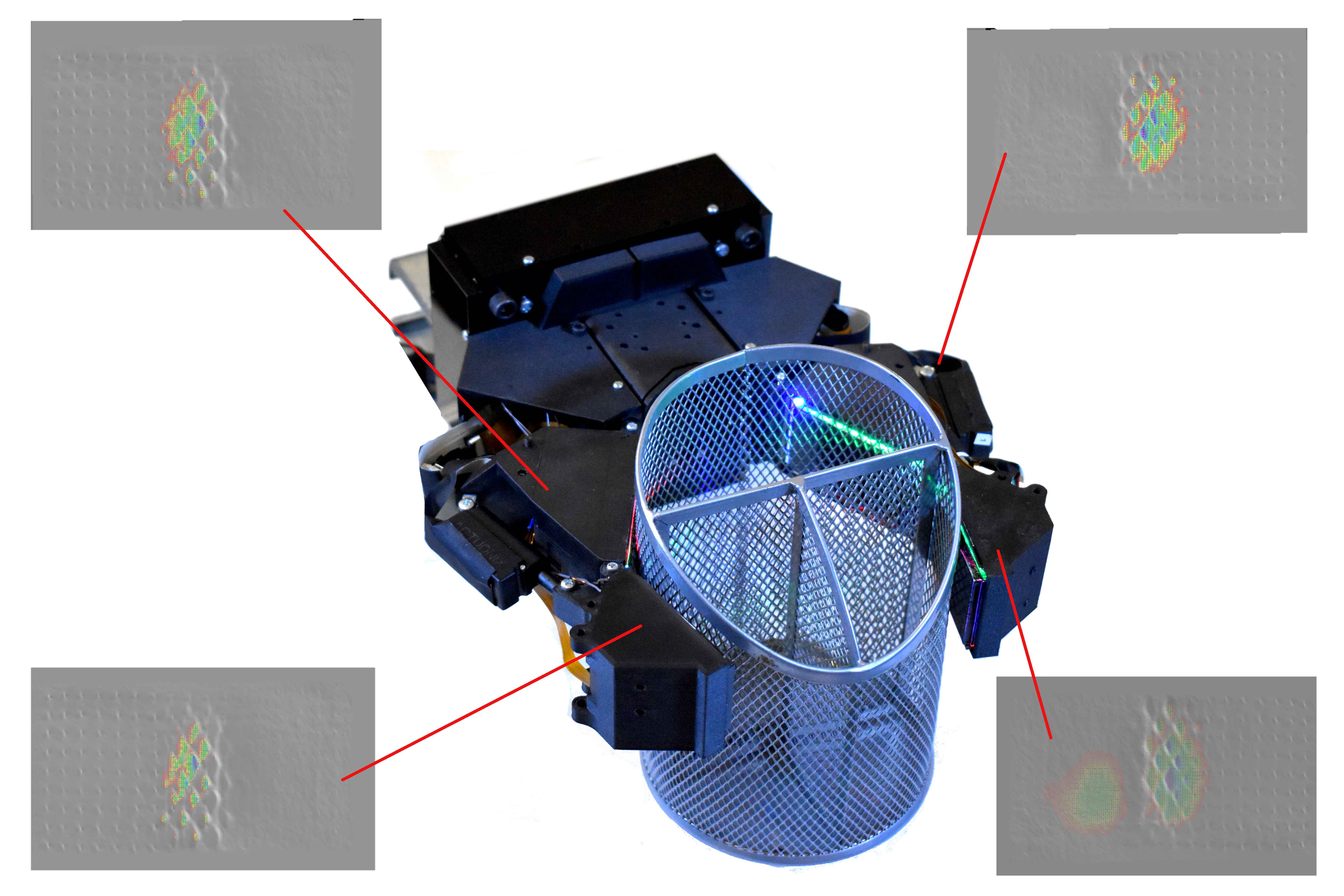}
    \caption{The pointcloud generated when the hand grasps a mesh pen holder}
    \label{fig:pointcloud}
\end{figure}

\section{Low Level Design}
The implementation of the high level design described in previous section is detailed below

\subsection{Tactile sensing using Gelsight}

The tactile sensing capabilities of the hand are designed around the Gelsight optical tactile sensing technology. The Gelsight tactile sensor consists of a layer of silicon based
elastomer which has a reflective layer and an array of markers.
A miniature camera, along with a directional lighting system is placed behind the silicone layer, inside
each phalanx. The lighting system illuminates the reflective side of the elastomer with primary colors from three directions. The 3D shape of the contact surface of the object is reconstructed
using photometric stereo technique\cite{Johnson:2011:MCU:1964921.1964941}. The 3D shape is computed from image gradients using Poisson image reconstruction.

The silicon layer also embeds optical markers, which are laser etched and painted. The proximal phalanx has an array of 9x17 markers, spaced at horizontal and vertical distance of 3mm, whereas the  distal finger has an array of 9x13 markers. The optical markers are tracked to create a vector field which corresponds to the deformation/displacement of the elastomer which is an indirect indication of the dynamic forces acting on the grasped object. The vector filed can be analysed to make sense of the normal, shear and torsional forces and slip acting on the surface \cite{8202149} \cite{8770057}. The following criterion is used to distinguish the type of force acting on it:
\newline
\begin{itemize}
  \item Shear force: if the magnitude of the vector summation, $| \sum_{i=1}^{N-1} \vec{v}_i | $ gives a larger value, than the curl listed below
  \item Torsional force: if any of the curl of the vector field at the points, $ \forall _i \in _{1..N} \nabla \times \vec{v}_i $  has a large value
\end{itemize}

where N is the number of the dots that are being tracked.
\newline

For marker tracking algorithm, first, we calculate the center of the markers with color filters and blob detection. Then we apply the marker matching to match the marker array. For marker matching, the straight-forward method is to compare the current frame with the previous frame and match with the nearest one for each marker. But this will cause hysteresis. When one frame went wrong, the following frames would keep the mismatching errors. 

Instead, we propose to improve the matching process by making use of the fact that we know the marker's layout when we fabricate the gel and markers. The layout includes the number of rows and columns of the markers and their interval distance. With this information, then we apply depth-first search (DFS) with pruning to maximize the smoothness of the marker flow. When missing or abundant markers are detected, the algorithm will also infer which ones are not detected correctly and interpolate to maximize the smoothness of the flow. Because we know the fabrication layout, many pruning techniques can be used to accelerate the optimization. For example, with the designed marker interval and gel elasticity, the marker at position (r, c), row r and column c, will stay on the left of the marker at position (r, c+1), likewise, will stay on the top of the marker at position (r+1, c), etc. By this way, we can get the matching process working with only the current frame and the initial frame. This solves the hysteresis problem since it doesn't depend on the previous history. For run-time speed, it's related to how many markers are not detected correctly. In our case, it can handle up to 4 markers not detected correctly by 30 Hz, and it can reach 1k Hz when all the markers are correctly detected.

The cameras used are Raspberry Pi V1.3 Camera, based on OV5647 sensor. The particular model is chosen because of its compact size, large number of lenses available and cheap cost. The lens with a wide field of view of 160\degree  is used. The wide field of view makes it possible for the camera to be placed closer to the sensing surface, thereby allowing for a thinner profile. For illuminating the elastomer membrane, an array of RGB LEDs are used. Unlike the fixed color LEDs used in previous Gelsight designs, networked and programmable LEDs are used, which would allow for precise and easy tuning of individual color channel intensities.
The cameras on each phalanx operate at 320x240 resolution at 90 frames per second and has a measured latency of 70 ms. The image from the four cameras is captured by an image acquisition system based on four Raspberry Pis and then streamed to the processing PC, where the 3D surface topology is reconstructed and the markers tracked in parallel.

\begin{figure}[htp]
    \centering
    \includegraphics[width=8cm]{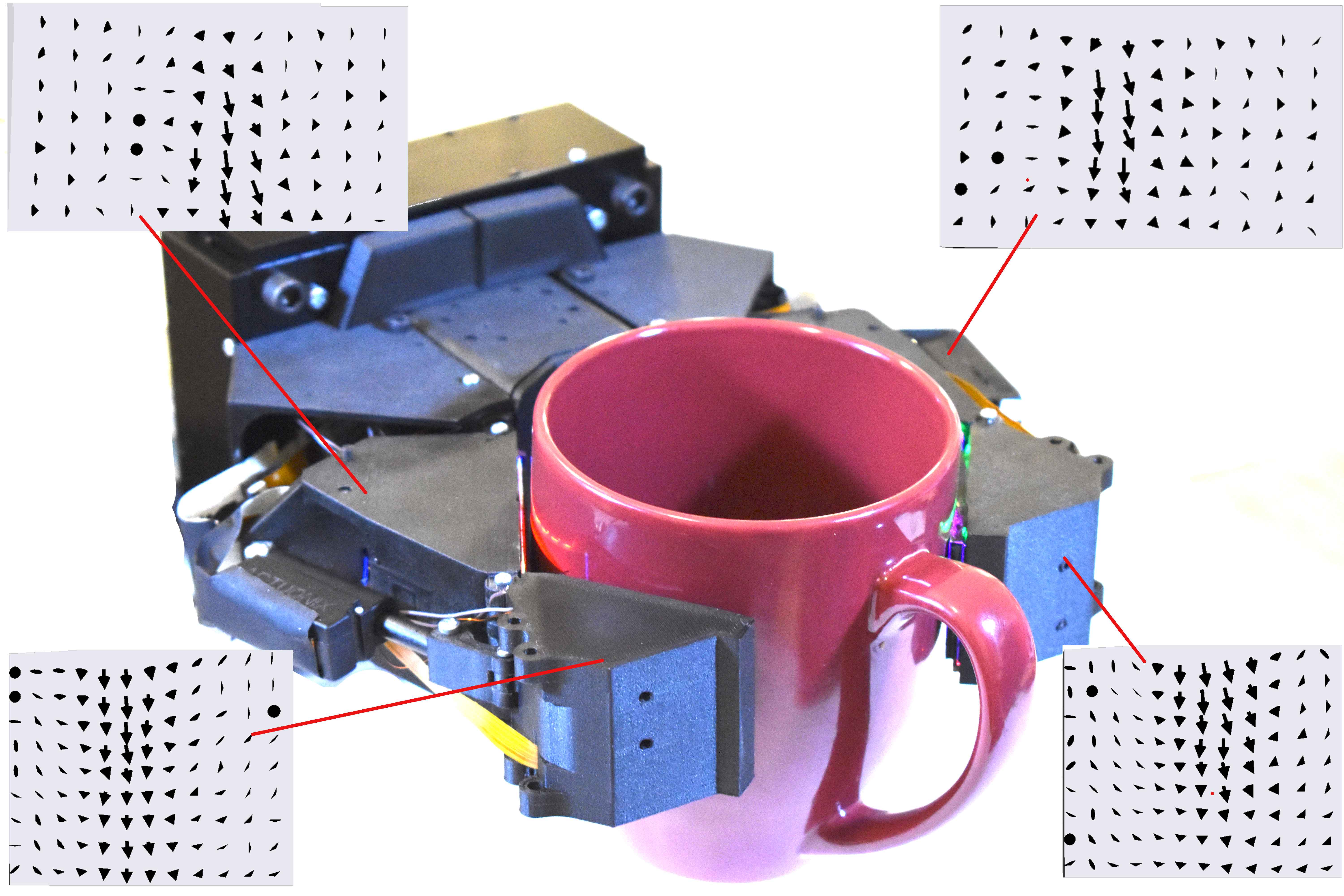}
    \caption{The vector field representing the displacement caused by the weight of the grasped object}
    \label{fig:galaxy}
\end{figure}

\begin{figure*} [t]
  \includegraphics[width=\textwidth,height=3cm]{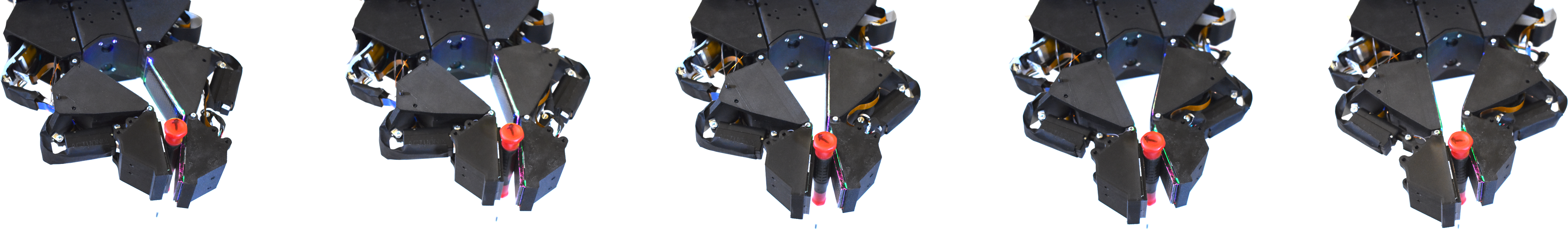}
  \caption{Demonstration of the in-hand manipulation capabilities of the hand, in which a screw driver is rolled counterclockwise from left to right.}
\end{figure*}
\subsection{Actuation and Proprioception}

The gripper consists of two fingers, each of which are made
of two phalanges that are fully actuated. The fully actuated two
degree of freedom fingers can operate in position
control as well as force control. It ensures that the gripper can
adapt to differently shaped objects as
well as perform reactive force control to ensure a secure grip,
especially during highly dynamic manipulator motions. 
The proximal phalanx has a range of motion of 95\degree, while the distal phalanx has a range of 150\degree.

Each phalanx of the finger is actuated using Linear DC motors with
built in position encoders.  The motors have an associated miniature lead screw for gear reduction. These linear actuators
are chosen owing to their high force to size ratio. The linear motors drive the
fingers through a linkage mechanism which, unlike the tendon
driven mechanisms, does not have the elastic drawback, or the
limited force, which are often limitations of tendon driven grippers. 

All the actuators used in the hand has positional encoders which are used for the closed loop position control of the hand as well as for the proprioception  of the hand. The encoder data is used to combine and project the data from the individual Gelsight sensors to a common frame of reference.

\subsection{Vibration and Acoustic Sensing}
One of the distal phalanx of the hand is equipped with a 3 axis MEMS accelerometer while the other one is equipped with a microphone. Both of them, augments each other and helps analyse the vibratory signals and the interactions with external world . It can be used to gather
information about the vibration generated when the finger moves
along a surface of an object. It can also be used to detect the
interaction of grasped object with external objects, like a contact with environment or even to detect the object slip. Humans detects object slip using micro vibrations in the skin \cite{99981}. The data from the accelerometer is sampled at 1 KHz. This ensures that the dominant frequency components that can be sensed by the human finger could be  extracted later by frequency domain analysis.

The gripper can also sense the auditory signals generated during the grasping or in-hand manipulation of objects and propagated along the hand structure. It augments the vibration sensing using accelerometer. To sense the sound signals, a miniature microphone is used, which can captures audio at 48 kHz sampling rate.The auditory signals can be very helpful when actions requiring contact with the environment are required. The dominant frequency component of the acoustic signal is determined using Fast Fourier Transform and is found to be quite reliable to detect/classify certain events. An experiment is designed to let the robot place a cup on a table and use the acoustic data to detect the contact. The data is shown as in table.

\begin{table}[h!]
\centering
\begin{tabular}{ |c|c|c| } 
 \hline
  Object & Surface & Frequency (Hz) \\ [0.5ex] 
 \hline\hline
   & Wood & 690  \\ 
 Glass & Marble & 1200 \\ 
  & Steel & 517 \\ 
 \hline
    & Wood & 345  \\ 
 Plastic & Marble & 520 \\ 
  & Steel & 173 \\ 
 \hline
    & Wood & 500  \\ 
 Paper & Marble & 710 \\ 
  & Steel & 375 \\ 
 \hline
\end{tabular}
\caption{Dominant sound frequency when placing a cup made of different objects on different surfaces }
\label{table:1}
\end{table}

\subsection{Imaging Sensors}
The palm of the hand is equipped with an Infra Red and a visible spectrum RGB imaging sensors, which can help in pre grasp planning and decision making. The IR imaging system is based on the MLX90640 IR array. It has a resolution of 32x24 with a wild field of view of 110x75 degrees and is sampled at 16Hz. It can be used to estimate the temperature of an object prior to grasping  or temperature distribution of an object prior to grasping via non-contact method with an accuracy of $ \pm $  1.5 \degree C. Thermal images  can also be helpful to train the robot to learn grasp poses from human demonstration \cite{Brahmbhatt_2019_CVPR}.

\begin{figure}[htp]
    \centering
    \includegraphics[width=8cm]{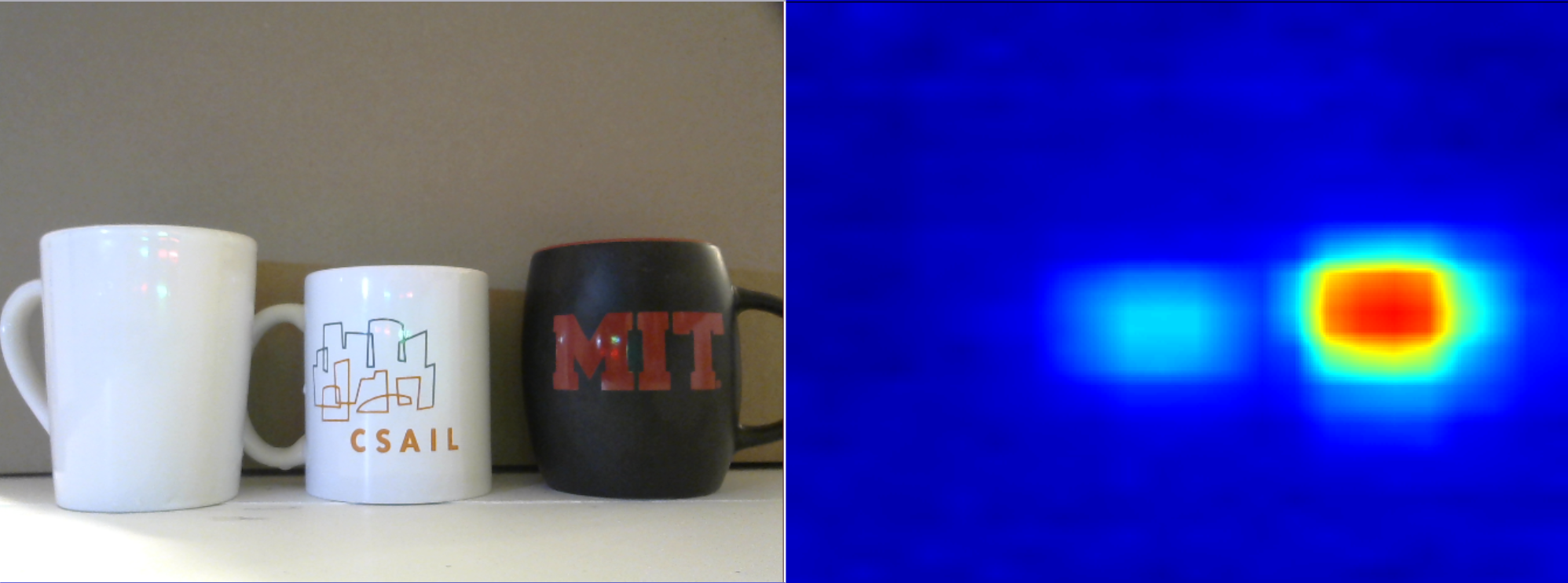}
    \caption{Image from the in-hand palm camera and the corresponding thermal camera. The left cup is filled with water at room temperature, middle cup with warm water at 50\degree C and right cup with boiling water at 90\degree C}
    \label{fig:galaxy2}
\end{figure}

The palm also has an RGB camera which streams video of 640x480 resolution at 30 fps. The visual information of the area in between the fingers could be used for pre-grasp object identification.

\subsection{Fabrication}
The structural components of the hand are mostly fabricated using Fused Filament Fabrication (FFF) 3D printing, using engineering quality nylon infused with carbon fibre. This ensures that the hand is strong and rigid enough. Using off the shelf components and 3D printed the structural components ensures that the costs are kept low as well as ease of replication and faster prototyping.

The silicone elastomer is molded on top of a clear acrylic substrate and has  shore OO 50 hardness. It is made from a two part silicone (XP-565 from Silicones Inc), mixed in ratio of 18:1 ratio of parts A:B. After mixed thoroughly,it is kept in vacuum chamber to remove the bubbles.The silicone is poured slowly and evenly on the rectangular acrylic substrate. A thin film of transparent silicone adhesive (Sil-Poxy from Smooth-On) is applied on the face of acrylic before the silicone is poured to enhance the adhesion. The silicone is poured slowly and evenly on the rectangular acrylic substrate.It is poured until it forms a film of approximately 1.5 mm thickness and is held by surface tension and forms a bulge at the edges.It is then kept inside an oven for 10 minutes for the silicone to partially cure, after which the next layer of silicone is poured over it, such that the total thickness is around 2-2.5 mm. It is then cured by keeping inside an oven at 200F for about 40 minutes. The two stage process  helps in making a thicker silicon surface than that would have been possible with a single stage process. The thicker gel would allow for more deformation, more contact area with the object and would capture more tactile information. 
The cured silicone on top of the acrylic is then spray painted with gray colored silicone ink. After the paint dries, dots are etched on the surface using a laser-cutting/etching machine. This would leave holes on the paint layer, which is then painted with black colored silicone paint, thereby giving black dots on a contrasting gray background.

The base of the hand also houses the embedded electronics systems, mainly the motor controller, real-time CPU for motor control,sensor data acquisition, lighting control and the camera interface boards. 

\subsection{Software Architecture}
The software architecture of the hand can be divided into the real time embedded software running on the hand and  the computational software running on an external PC. The embedded real time software does the closed loop position and force control of the actuators, sampling of the sensors including accelerometer, IR array, microphone etc and streams it to the controlling PC .
The data from all the sensors are streamed over a Gigabit Ethernet Link. The software running on the computer is built on top of the  Robot Operating System(ROS) architecture. There are ROS nodes corresponding to each of the sensor, which interfaces with them and publishes the data stream as corresponding ROS messages. These messages are then subscribed by nodes which performs 3D reconstruction using photometric stereo, optical marker tracking etc. This ensures that the system is modular and flexible to customization. The display tool in ROS - rviz, is used to visualize all the sensor data.

 \section{Challenges, Lessons learned and Future directions}
 There were numerous challenges encountered while trying to design a hand which has all the sensing modalities and actuation 
 \newline
\begin{itemize}
\item \textbf{Compactness}  Squeezing the actuation and the optical sensors into a compact size turned out to be challenging since the optical Gelsight sensors are inherently bulky and the actuation mechanisms must not get in its way. The problem in tackled by using camera with a wide field of view, such that it can be placed closer to the membrane and opting for a linear actuator which has a higher force per volume. The actuator is placed external to the bulk of the finger phalanx.

\item \textbf{Contact Surface} - The contact surface area of the flat shaped membrane surface is low with common objects, which are often curved. Grasping objects which are tapered in shape resulted in contacts only at the widest point. This limits the amount of useful data that can be acquired. It may also increase the chance of having an unstable grasp. Contact surface designs with curved or convex membranes paired with thicker soft membranes may have to be explored in future, which could help reduce this shortcoming

\item \textbf{Signal cross talk and interference} - In the current version , the signals from each camera are carried over Flexible Flat Cables (FFC) to the corresponding Raspberry Pi. These cables are normally unsheilded and running them in parallel along the length of about 2 meter of a typical robot manipulator can give rise to serious crosstalk and corruption of the data. Shielding the cables and maintaining distance between each cable line solved the issue. In future iterations, the image acquisition systems needs to be miniaturized and shifted to be in close proximity with the sensors.

\item  \textbf{Bandwidth  and computational limitations} -  Streaming images from four Gelsight sensor cameras and an additional palm camera requires serious data bandwidth, that eventually required a Gigabit Ethernet connection to fetch frames with an acceptable latency.   In addition, the Poisson solver for reconstructing 3D has a high computational requirement, which required the usage of a powerful PC. It would be much efficient if the computation is offloaded to some dedicated hardware, like an FPGA, inside the hand itself, so that both the challenges of bandwidth and processing power can be handled

\item \textbf{Dexterity} -  The two fingers which has 2 DOF and are fully actuated, had much better dexterity and in hand manipulation capabilities than an underactuated hand. It was also observed that adding a one more joint, preferably actuated, would help to have much better dexterity, and make independent positioning and orienting possible, which is not optimal with the current design. 
\end{itemize}
 
\section{Conclusion and Future Directions}

An integrated adaptive gripper with multiple sensing modalities was developed, each of them tested and studied. The main contributions of the work are integrating actuation, and multiple sensing technologies, especially multiple Gelsight based sensors to a compact form factor in a single hand.The fully actuated hand helps in achieving various grasp poses and may also be used for tasks like active tactile exploration, in hand manipulation etc. Fig 5 shows an experiment in which the dexterity of the hand is used to re-orient an object. The screw driver is rotated in a counterclockwise direction from left to right. The kinematics of the hand can be calculated on the fly, based on the position of the object, which is sensed by the Gelsight sensors. The compact form factor enables the hand to navigate in and out of tight spaces in cluttered environments. The usage of a multi sensory approach to robot manipulation and grasping can increase the reliability as well as bring in new unexplored capabilities.

In addition to the development of the gripper, novel
algorithms must be developed to fuse the data generated by the
multitude of sensors to bring new capabilities as well as improve
the reliability of robot manipulation

\bibliographystyle{./bibliography/IEEEtran}
\bibliography{./bibliography/IEEEabrv,./bibliography/IEEEexample}

\vspace{12pt}

\end{document}